# Semantic Knowledge Technologies: what the Semantic Web lost sight of, and what it never had

**Achille Zappa**

Glycan and Life Systems Integration Center (GaLSIC), Soka University, Hachioji, Japan

---

## Abstract

The Semantic Web set out to give information a machine-interpretable form so that software could integrate and reason over it. Its standards became scientific knowledge infrastructure, but the machine competence it promised did not follow, and the systems now answering questions over scientific knowledge are language models holding no inspectable account of what they know. This paper argues the original goal was right and the technical programme incomplete, states what is missing, and names the extended programme Semantic Knowledge Technologies: the same technical core carried out of its web-publishing origin and applied to knowledge wherever held. The diagnosis is that the standards formalised truth while omitting three things: the conditions under which a claim holds, the operations its terms permit, and any account of what a base covers. Without conditions, contradiction and applicability cannot be judged; without operational grounding, holding a statement confers no ability; without declared coverage, a system cannot recognise the boundary of its own content, which under the open-world assumption cannot be inferred. The paper fixes the word *understanding* to five measurable tests (check, connect, derive, act, delimit) and sets out a seven-layer architecture in which the first three layers are enabling and the rest the cognitive capabilities they make possible. It then defines three terms the programme implies: **Large Knowledge Model**, a model whose unit of output is a reference to an addressable claim, not a token; **SLKM**, the knowledge base an agent builds for itself from declared sources; and **Semantic Artificial General Intelligence**, stated as a falsifiable position about necessary conditions, not a system. A graded ladder replaces the untestable word *general*. It is offered as a research agenda, with its weakest points and refutation condition named.

## 1. What this paper is about

The Semantic Web programme set out to give information a form that machines could interpret, so that software could integrate it and reason over it [1]. It produced the Resource Description Framework (RDF), the Web Ontology Language (OWL), the SPARQL query language, and later the Shapes Constraint Language (SHACL) and the Provenance Ontology (PROV-O) [2]. These standards are in daily use as scientific knowledge infrastructure. The machine competence the programme promised did not follow. What arrived instead, twenty years later, was a different technology: language models that answer fluently from statistical regularities in text, with no inspectable account of what they hold.

This paper takes the position that the original goal was correct and the technical programme was incomplete, and states what has to be added. The goal is knowledge represented so that machines can reason over it, learn from it, discover with it, and present results whose basis a person can inspect and use. I use **Semantic Knowledge Technologies** for the extended programme, and **SemanticKnowledge** for the specific profile and architecture defined here. The word *Web* is dropped because nothing in the programme depends on web publication: the same requirements apply to institutional knowledge bases, curated databases, laboratory records and knowledge extracted from literature.

One term is used throughout in a strong sense. **Knowledge**, here, means information recorded so that its meaning, its conditions of application, its origin, its reliability and the operations that apply to it are all explicit and machine-interpretable. That is stronger than data, stronger than structured data, and stronger than what the current standards require.

## 2. What understanding means in this programme

Claims about machine understanding are usually either grandiose or empty. This programme fixes the word to five tests, each of which can be measured. A machine understands a statement to the degree that it can:

1. **Check** it, against declared constraints and against what it already holds, and report a verdict with a reason.

2. **Connect** it, resolving its terms to entities and relations already held.
3. **Derive** from it, producing consequences with the inference method recorded and the derivation traceable to premises.
4. **Act** on it, by invoking the computations to which its terms are bound.
5. **Delimit** it, by reporting the coverage, currency and confidence of what it holds, and refusing what falls outside.

Understanding in this sense is not a component to be built. It is a property that the architecture below confers, and the five tests are how the architecture is evaluated. No claim is made that a system passing all five is intelligent in any wider sense, and the programme does not require such a claim in order to be worth pursuing.

## 3. Why the Semantic Web could not produce reasoning

The common account of the failure is sociological: annotation was expensive, ontologies did not converge, the incentives were wrong. All true, and it explains slow adoption rather than absent capability. The technical account is more useful, and it is this: **the standards formalised truth and left out conditions and competence.**

**The first omission concerns claims.** A triple asserts that something is the case. It carries no time, no scope, no conditions of application, no method, no evidence, and no degree. OWL then tells you what follows from a set of such assertions. But a claim in any empirical field holds under conditions, was established by a method, is supported by evidence of some strength, and may be contested. Reduce it to a context-free assertion and the information a reasoner would need in order to reason responsibly has already been discarded. Two claims cannot be recognised as contradictory when the conditions that would distinguish them are absent, and applicability cannot be judged when scope was never recorded.

**The second omission concerns terms.** In RDF and OWL a term means whatever other terms say about it. The semantics is denotational: it fixes what expressions refer to and what entailments hold, and it connects nothing to computation. A machine holding a statement about a glycan structure can match the symbol and cannot convert, compare, resolve or verify it, because nothing in the standards records that such operations exist. Meaning had no operational component, so holding knowledge conferred no ability.

**The third omission concerns the system itself.** A knowledge base has no standard way to state what it covers. Under RDF's open-world assumption the absence of a statement is not evidence of anything, so a client cannot distinguish a gap in the data from a gap in the world, and a system cannot recognise the boundary of its own content. Existing dataset description records where data is and what a service supports technically, not what its holder can competently answer.

These three omissions are why adding more triples was never going to produce reasoning, and they are the starting point of the agenda below.

## 4. The architecture: seven layers in three groups

Layers 1 to 3 are **enabling**: they make everything above them trustworthy, and they are where the standards work lies. Layers 4 to 6 are the **cognitive capabilities** the programme exists to reach. Layer 7 is **delivery**, where results become usable.

Each layer is set out as what it provides, its mechanism and technology, the relevant prior art, the hard part, and its status. The status lines matter: this is an agenda, and it should be clear at every point which parts exist, which are designed but unbuilt, and which are open research.

### Layer 1. Qualified claims

**Provides.** The unit of content is not a bare triple but a claim carrying the conditions under which it holds, the provenance of its assertion, an evidence category, and a lifecycle status (candidate, admitted, rejected, superseded, retracted, contested). This is what makes applicability judgements and contradiction detection possible at all.

**A worked example.** As a bare triple, a glycosylation record asserts that a protein carries a glycan at a position, and nothing more:

```
uniprot:P02724  hasGlycosylationSite  62 .
```

As a qualified claim, the same content carries the organism and tissue in which it was observed, the experimental method, the source resource and its release identifier, an evidence category recording whether it was curated, computed or extracted from text, the provenance of any extraction including the model version and source passage, and a lifecycle status:

```
:claim-8831 {
  uniprot:P02724  hasGlycosylationSite  62 .
}
:claim-8831 prov:wasDerivedFrom      glycosmos:release-2026-03 ;
            :organism                ncbitaxon:9606 ;
            :tissue                  uberon:0002371 ;
            :method                  :mass-spectrometry ;
            :evidenceCategory        :curated ;
            :status                  :admitted .
```

A second record from another resource asserting a different glycan at the same position is then either a genuine contradiction, if the conditions agree, or two observations under different conditions, if they do not. The bare triples cannot be told apart; the qualified claims can. Every capability in the layers above depends on that distinction.

**Mechanism.** Statement-level annotation using RDF 1.2 triple terms and the `rdf:reifies` property, which make statements about statements native to the data model, with named graphs as the portable fallback in current software. Provenance in PROV-O. Evidence as a small controlled vocabulary tied to how the claim was obtained: curated by a resource, derived by a declared computation, extracted from text by a named model version, or asserted by a human. Conditions expressed through a compact schema defined per assertion type and per domain.

**Prior art.** Nanopublications established packaging an assertion with its provenance and attribution as a single citable unit [8], and were later given a decentralised publishing and verification infrastructure [9]; micropublications developed the evidence and argumentation side, modelling claims together with the evidence and discussion supporting or opposing them [3]. This layer follows both and adds validation status and lifecycle.

**Hard part.** Choosing the condition dimensions. No universal theory of context is attempted here; the design commitment is a small, explicit, domain-parameterised condition schema per assertion type, accepting that it will be revised in use. If the chosen dimensions are wrong, contradiction detection under-reports and the layer delivers little, which is the main way this part of the programme could fail.

**Status.** Mechanisms exist; the condition schema and the profile do not. The historical obstacle was the cost of authoring claims at this level of detail, and assisted extraction has reduced that cost substantially.

**Layer 2. Executable terms**

**Provides.** Terms bound to the operations that apply to their instances, so that holding a statement confers the ability to do something with it. This gives meaning an operational component alongside its denotational one.

**Mechanism.** A binding vocabulary recording, for a vocabulary term, the operations available on its instances: the operation type (validate, convert, resolve, compute, retrieve), an invocation description sufficient to call it, the inputs and outputs typed against the vocabulary, and the provenance to attach to results. Written independently of any particular invocation protocol, with protocol adapters kept separate. In glycoscience the bindings wrap published tools: structure notation conversion and comparison, resolution of structures to repository accessions, checking of reported glycosylation sites against protein records, retrieval of biosynthetic and disease context.

**Prior art.** OWL-S and SAWSDL described services in detail and were not adopted; SADI bound services to ontology classes so that discovery and pipelining could be driven by the semantics of their inputs and outputs [4]; EDAM and the bio.tools registry annotate bioinformatics operations, data types and formats at community scale [5, 6]. The difference proposed here is scope and purpose: term-level, minimal, aimed at invocation by an agent rather than automated composition.

**Hard part.** Scope discipline. This is exactly where earlier efforts became unusable. The rule adopted is that a term is bound only when a concrete task requires it, and that operations are described rather than modelled.

**Status.** To be built. An ecosystem of tool-invocation interfaces for agents now exists, which is the infrastructure the earlier attempts lacked.

**Layer 3. Declared coverage**

**Provides.** A machine-readable account of what the base holds: which regions of content, from which sources at which release, at what currency, and which regions are asserted complete for a stated population. This is what allows a system to recognise a question it cannot answer, and a client to route a question elsewhere.

**Mechanism.** A coverage vocabulary defining regions by entity types and predicates, with source, release identifier, currency date and an explicit completeness claim where one can honestly be made; plus a procedure mapping a question to the regions its terms require, and a refusal that names the missing region. Generated from the store, the constraints and the binding registry on every build so that the description cannot drift from the content.

**Prior art.** Dataset description practice records what a dataset contains and what a service supports: VoID, SPARQL service descriptions, and the dataset description profile developed in the health care and life sciences community, which covers description, identification, attribution, versioning, provenance and content summarisation [10]. This layer extends that toward what a holder can competently answer.

**Hard part.** The open-world assumption means non-coverage cannot be inferred, only asserted, so coverage is a maintainer's declaration over bounded regions. The open empirical question is how coarse a region can be before refusal decisions become unreliable.

**Status.** To be built. This is the layer with the least prior art and the sharpest contrast with language models, which cannot hold such a description by construction.

### Layer 4. Inference

**Provides.** Derivation of new statements from held ones by several methods, each labelled, so that a consumer can always distinguish what was checked from what was estimated or guessed.

**Mechanism.** Deduction where the axioms genuinely hold, using tractable OWL 2 profiles and rules expressed with SHACL-AF or SPARQL constructs, for classification, constraint entailment and transitive closure. Induction over graph structure using knowledge graph embedding and link prediction, for ranking plausible completions. Abduction and analogy, where a language model proposes candidate explanations or relations that are then subject to the same admission gate as any other content. Every derived statement carries its method, its confidence and a trace to its premises, and derived statements are stored as claims with provenance recording the derivation rather than being silently merged with asserted content.

**Hard part.** Combining methods without producing a system whose outputs cannot be interpreted. The labelling discipline is the answer, and the honest limitation is that no principled semantics exists for combining a deductive verdict with an embedding score; the profile records both and does not pretend to fuse them.

**Status.** Components mature and separately available. The orchestration and the labelling discipline are the contribution.

**Layer 5. Learning and memory dynamics**

**Provides.** Knowledge that grows and corrects itself, instead of a static release. Also the substrate on which learned models can be trained or grounded.

**Mechanism.** Acquisition through mappings and assisted extraction; consolidation, meaning the integration of repeated or overlapping observations into stable claims with accumulated evidence; revision when contradictions are detected under shared conditions; supersession and retraction as recorded operations rather than deletions; decay of confidence as evidence ages, with the ageing rule stated per assertion type. A feedback loop in which validation rejections, refused questions and failed queries are treated as signals that improve extraction, mappings, constraints and vocabulary. Learned representations recomputed as content changes.

**An analogy, used with care.** Consolidation, re-linking and decay in this layer play the role that synaptic plasticity and memory consolidation play in a brain: the graph does not only grow, it rearranges, strengthening what evidence reinforces and letting go of what nothing supports. The analogy communicates the intent well and is used in that spirit; the mechanisms are graph operations with recorded provenance, and no neuroscientific claim is made or needed.

**Hard part.** Belief revision at scale has no settled practical solution, and consolidation rules are domain-specific. The scope commitment is to specify the operations and record them, not to solve revision in general.

**Status.** To be built. Acquisition and revision are tractable now; consolidation policy, decay and the full feedback loop are the least mature part of the programme and should not be promised in a short project.

**Layer 6. Discovery**

**Provides.** Output that is not retrieval: identified gaps, surfaced contradictions, and candidate hypotheses with their support.

**Mechanism.** Gap detection reads the coverage declaration and reports regions that are thin, stale, or never examined, including combinations of entity types that no source addresses; this is only possible because coverage was declared, since an undeclared base cannot distinguish an empty region from an absent one. Contradiction surfacing compares claims that share conditions and disagree, which is only meaningful because conditions were recorded. Hypothesis generation combines link prediction over the graph with model-proposed candidates, and every candidate passes the admission gate before it is stored, with its status marked as proposed rather than admitted.

**Hard part.** Precision. A discovery layer that emits many weak candidates is worse than none, because it consumes expert attention. Ranking and a strict admission gate are the controls, and the measure of success is expert-judged yield per candidate reviewed, not the number of candidates.

**Status.** To be built. This is where the enabling layers pay off, and the argument for treating the whole as one system rather than as three separate good practices.

### Layer 7. Presentation and exploitation

**Provides.** Results that a person or a downstream system can act on: an answer together with the claims that support it, the conditions under which it holds, what was checked as opposed to inferred, the alternatives considered, the confidence, and what the system could not cover.

**Mechanism.** Rendering of derivation traces into readable justifications; a language layer translating questions into queries and results into prose, constrained to admitted content and required to refuse visibly when coverage does not support an answer; structured views over claims and their evidence, including comparison of conflicting claims side by side with their conditions; interactive refinement, where a user narrows conditions or expands a region and sees the answer change; and programmatic access so that other systems consume claims with their qualifications intact rather than as flat records. Every presented element remains linked to the claim identifiers behind it.

**Hard part.** Presenting uncertainty and scope without either overwhelming the reader or flattening the qualifications that the lower layers took care to record. This is a design problem, and it is where the programme's value becomes visible or fails to.

**Status.** To be built on top of the rest. The requirement is that nothing is presented that cannot be traced back.

## 5. How the layers depend on one another

The dependencies are what make this one architecture rather than a list. Inference over unqualified claims produces confident nonsense, because contradiction and applicability are undecidable without conditions. Action on ungrounded terms produces results that cannot be verified. Learning without declared limits produces drift that nobody detects. Discovery without coverage declarations cannot distinguish an empty region from an unexamined one, and without conditions cannot tell a real contradiction from missing context. Presentation without provenance and traces reduces to assertion, which is the failure mode of current systems.

Conversely, the enabling layers are not ends in themselves: a base that qualifies its claims, grounds its terms and declares its limits, and then only answers lookup queries, has paid a cost for nothing.

**Relation to FAIR.** The FAIR principles [7] ask that data be findable, accessible, interoperable and reusable, with explicit emphasis on machine-actionability. That programme concerns getting to the data and being able to process it. The requirements set out here concern what a machine must be able to do with the content once it has it: judge its applicability, act on its terms, and know where its knowledge ends. The two are complementary, and neither implies the other.

## 6. How it is used

Four modes, and the architecture should be judged in all four.

A **machine agent** queries the base, invokes bound operations, receives labelled derivations, and refuses correctly when coverage does not extend to the question.

A **scientist** asks a question in ordinary language and receives an answer with its basis, its conditions and its gaps, and can move from the rendered answer to the underlying claims.

**Independent knowledge bases** exchange claims with conditions, provenance and evidence attached, so that each can decide by its own policy what to admit, which is what makes federation between institutions meaningful rather than a schema-matching exercise.

**Learned models** are trained or grounded on curated claims instead of scraped text, with provenance available for filtering and weighting. A fifth mode combines these: an autonomous agent constructs and maintains its own base as its memory. Fed the endpoints of declared sources, it studies them, grows the base it answers from, and enriches the sources in return with the annotations its study produces.

## 7. Hypotheses

The programme is stated as claims that can fail.

*On the enabling layers.* Representing claims with conditions and evidence enables correct answers to applicability and contradiction questions that the same content as bare triples cannot support. Binding terms to operations enables multi-step tasks whose success in unbound systems degrades as the number of steps grows. Declared coverage enables correct refusal and correct routing, which neither additional content nor a larger model provides.

*On the cognitive layers.* Labelled multi-method inference gives higher usable accuracy than deduction alone, without the loss of interpretability that unlabelled mixing causes. A maintained base stays correct longer under a stream of new evidence than a static curated release. Discovery over declared coverage and recorded conditions yields expert-confirmed gaps and contradictions at a rate that justifies the review effort.

*On delivery.* Results carrying their basis and limits change what a user does with them, measured by correct acceptance and correct rejection of system output.

*At system level.* A machine restricted to this architecture answers domain questions with higher correctness and traceability, and fails correctly rather than confabulating, compared with the same model given the same content as text. The cost of construction and maintenance is measurable and worth reporting whichever way the comparison falls.

## 8. Scope

**Inside the programme:** the representation profile; vocabularies for bindings and for coverage; assisted authoring of qualified claims; admission and validation; inference orchestration and its labelling; memory operations; discovery functions; presentation requirements; the evaluation method; and domain testbeds.

**Outside, and stated as outside** so that the programme is not read as claiming them: any new logic or ontology language; a general theory of context; a formal probabilistic semantics for combining confidence; general-purpose belief revision; artificial general intelligence, and any claim that this architecture constitutes or leads to it; and interface design research beyond the requirement that presented content be traceable.

The level of the work is infrastructure and method: a profile over existing standards, two candidate vocabularies, an architecture, and a way of measuring whether it delivers.

## 9. Sequence

The programme is too large for one project, and the order matters because of the dependencies in section 5.

The first step takes layer 1 (qualified claims), layer 3 (declared coverage), the acquisition and revision parts of layer 5, and the discrepancy-detection part of layer 6, and puts them inside an autonomous learning system in a single domain where the semantic resources are mature. Concretely: two systems study a domain's knowledge bases, the ontologies that describe them, the mappings between vocabularies, and the literature; they admit only validated claims, marking anything imported from an existing resource as such so that what they establish themselves stays distinguishable; they maintain their own coverage accounts; they exchange and re-validate claims with each other; and they are examined longitudinally by domain specialists, with the language model replaced partway through to locate where the acquired expertise resides. What they learn is applied outward as annotation of literature and as proposed corrections to vocabulary mappings, which is a first, bounded instance of layer 7.

What follows, in rough order of dependence: executable bindings for terms (layer 2); the full inference portfolio with its labelling discipline; consolidation and decay beyond acquisition and revision; hypothesis generation with expert-judged yield; presentation and exploitation as a designed layer rather than a minimum; exchange between independently maintained bases across institutions; and a second and third domain, which is the only way to test whether the profile and vocabularies are domain-independent as intended. A model trained and constrained in the way described in section 11.1 belongs at the end of this sequence, because it requires the substrate to exist at scale first.

Mapping the sequence onto the generality levels of section 11.2: the first project sits at G0, the second and third domains are the test of G1, and exchange between bases is the test of G2.

## 10. Risks and honest weaknesses

The parts are not individually new. Statement-level provenance, service binding, dataset description, neuro-symbolic inference and knowledge graph maintenance all exist as research lines. The claim made here is about the combination and about the dependencies between the parts, and that claim has to be stated plainly rather than dressed as novelty of components. A reader who says this is nanopublications plus tool use plus a coverage vocabulary is right about the ingredients and has not yet engaged with the argument that competence requires all of them together.

The condition schema of layer 1 is the main intellectual risk: if the dimensions chosen are wrong or too coarse, applicability and contradiction handling will not improve, and the first hypothesis fails. Scope discipline on layer 2 is the main engineering risk, with a clear historical precedent for failure. The granularity of coverage declarations in layer 3 is unexplored and may turn out to require impractical detail. Layer 5 is immature and should not be promised in short projects. Layer 6 will be judged on precision, and a discovery layer that wastes expert attention is worse than no discovery layer. The programme depends on standards still in progress, so portable fallbacks are specified for every mechanism that relies on them.

A final weakness worth stating: the whole programme presumes that traceable, conditioned, refusable answers are worth their construction cost. That is an empirical question, and it is the one the first project is designed to answer.

## 11. Three terms, defined

Three terms are used in the development of this programme: Large Knowledge Model, Semantic Artificial General Intelligence, and SLKM. I introduce all three here. Left undefined they would be branding, and earlier drafts of this paper omitted them for that reason. Defined, they name three things the programme implies: an architecture, a position together with a class of systems, and an artefact. Each is given below with a definition precise enough to be argued with, and with its weakest point named.

### 11.1 Large Knowledge Model

A **Large Knowledge Model (LKM)** is a learned model that stands in a different relation to knowledge than a language model does, in three specific respects.

**First, substrate.** Its training and grounding material is a body of qualified claims as defined in layer 1, each carrying conditions, provenance, evidence category and status, rather than uncurated text. Provenance is available during training, so material can be filtered and weighted by evidence category rather than by scraping heuristics.

**Second, and this is the substantive difference, the unit of output.** A language model emits tokens, and the assertions a reader recovers from those tokens are not addressable objects: nothing in the output points at anything. An LKM emits references to addressable claims and to operations, and readable text is rendered from the resolved references. The consequence is a hard constraint rather than a tendency: every statement in an output is either resolvable to an admitted claim, resolvable to a recorded operation and its result, marked as a proposal awaiting validation, or absent because the model refused. Unattributable assertion becomes structurally unavailable, which is not something training objectives or guardrails can guarantee.

**Third, updatability and self-report.** Because what the model asserts is held in an external store with declared coverage, correcting knowledge means editing claims rather than retraining, and the model can report what it covers by reading layer 3 instead of estimating its own competence.

**What it requires:** layers 1 to 3 for the substrate, layer 4 for labelled derivation, layer 5 for the training and maintenance loop, layer 7 for rendering. **How to test it:** the five tests of section 2, plus the proportion of output statements that resolve to admitted claims or recorded operations, which should be one by construction and can be audited, and the correctness of coverage self-report. **What is speculative:** whether a model constrained to emit claim references retains enough fluency to be useful, and whether training on qualified claims at feasible scale is competitive with training on text at very large scale. Both are open, and the second is the harder objection. The phrase *large knowledge model* appears elsewhere with looser meanings; the definition above is specific to this programme.

### 11.2 Semantic Artificial General Intelligence

**Semantic Artificial General Intelligence (SAGI)** carries two related senses here, and keeping them apart keeps both usable.

In the **strong sense** it is not a system this programme proposes to build: it is a position about necessary conditions, stated below so that it can be attacked.

In the **operational sense**, a SAGI names a concrete class of systems, the semantic discerning agents: machines that are pointed at declared semantic sources (SPARQL endpoints, ontology repositories and their cross-ontology mappings, RDF datasets local or remote) and that autonomously explore, examine and study them; that learn the data, the structures, the models and the relations they find; and that build from this study their own knowledge base, their SLKM (section 11.3), which serves them as memory and context and which they continue to expand, reorganise and correct.

**The position.** General machine competence over knowledge requires an external, addressable body of claims that carry their conditions and provenance, terms bound to operations, declared coverage, labelled inference, and specified revision. The argument is that the five tests of section 2 cannot all be satisfied by a system whose knowledge exists only as fitted parameters: checking requires declared constraints, traceable derivation requires addressable premises, and delimiting requires a model of coverage that a parametric system has no place to keep. On this view, statistical learning supplies interpretation, proposal and rendering, and cannot supply the substrate; so any route to general competence passes through something with the properties described in this paper, whether or not it is called by this name.

**Generality, graded.** The word *general* is what makes such claims untestable, so the programme replaces it with a ladder of four levels, each with a test.

- **G0, single domain.** The five tests hold within one base's declared coverage.
- **G1, cross-domain.** The same profile and vocabularies carry to a second and third domain with only the bindings replaced, which tests whether the design was domain-independent as intended.
- **G2, cross-base.** Independent bases exchange qualified claims and competence is preserved end to end, tested by answering a question that requires composing two bases with provenance intact through the composition.
- **G3, open-ended.** The system extends its own coverage, acquiring new regions, revising its condition schemas, and proposing and validating new bindings without those being supplied by hand.

G0 to G2 are engineering milestones on the sequence in section 9. G3 is a research frontier, is unproven, and may not be reachable; nothing in this programme depends on it.

**The opposing view, stated fairly.** The scaling position holds that these properties can emerge from learning at sufficient scale, with verification behaviour trained in rather than imposed architecturally, and that explicit substrates carry maintenance costs that grow faster than their benefits. This position has strong empirical support in capability terms, and it is not obviously wrong about cost. The disagreement is therefore empirical, which is the useful outcome of stating it: **if a system without an external addressable store reliably passes all five tests at scale, including correct refusal at the boundary of its knowledge, the necessity claim above fails and should be abandoned.**

### 11.3 SLKM: the agent-built memory

An **SLKM (Semantic Learned Knowledge Memory)** is the knowledge base a SAGI agent constructs and maintains for itself: a SemanticKnowledge base in the sense of this paper, populated not by curators but by the agent's own study of declared sources, holding what the agent has learned, verified, annotated and derived, together with its coverage account. It is the instance-level counterpart of the LKM idea of section 11.1: where LKM names a model whose relation to knowledge runs through addressable claims, SLKM names the body of addressable claims a particular agent has accumulated: its memory, its context and, informally, its brain.

Three properties separate an SLKM from a cache or a scratchpad: every entry passed validation before admission and carries conditions, provenance and evidence; the structure reorganises as learning proceeds, through consolidation, re-linking, supersession and decay; and the whole is inspectable and correctable by a human at the level of individual claims.

**On the brain analogy.** My working intuition for the SLKM is explicitly analogical: claims and their typed links grow and rearrange the way neuronal connections do, consolidation strengthens what repeated evidence supports the way memory consolidation does, decay weakens what nothing reinforces, and the whole functions as reasoning-plus-memory the way an expert's long-term memory does. The analogy is a communication device and a source of design questions, and it earns its keep in both roles. It is not a neuroscientific claim: the mechanisms are graph operations with recorded provenance, and nothing in the programme depends on the analogy being more than an analogy.

**A note on use.** These three terms name real distinctions and are meant to be used: LKM for a model whose unit of output is a claim reference, SLKM for the knowledge base an agent builds for itself, SAGI for the class of systems that study declared semantic sources and for the position about what general competence requires. Where a definition can travel with them, as in this paper, they should be used. In method sections and specifications, where a reader meets the term without the definition, the underlying mechanisms should also be named directly (qualified claims, bindings, declared coverage, labelled inference, persistent semantic memory) so that the argument remains available to someone who rejects the vocabulary.

## 12. Naming

**SemanticKnowledge** is the proper name of the profile and architecture. **Semantic Knowledge Technologies (SKT)** is the family term, parallel to Semantic Web Technologies. A conformant base is a **SemanticKnowledge base**. LKM, SAGI and SLKM are used only as defined in section 11.

A note for literature search: the two-word phrase *semantic knowledge* has an established and unrelated meaning in cognitive psychology, referring to general world knowledge in human memory. That is a reason to prefer the one-word proper name and the family term in titles.